\documentclass[sigconf]{acmart}

\renewcommand\footnotetextcopyrightpermission[1]{} 
\usepackage{xcolor}
\usepackage{enumitem}
\usepackage{amsmath}
\usepackage{graphicx}
\usepackage{caption}
\usepackage{array}
\usepackage{booktabs}
\usepackage{natbib}
\usepackage{url}
\usepackage{dsfont}
\usepackage{gensymb}
\usepackage{subcaption}
\usepackage[ruled,vlined,linesnumbered]{algorithm2e}
\usepackage{algorithmicx}
\usepackage[noend]{algpseudocode}

\newcommand{\alg}{\textsc{HiPaR}}
\newcommand{\dataset}{$D$}
\newcommand{\mdataset}{D}
\newcommand{\parentrule}{$\top \Rightarrow y = f_{\top}(A'_{\textit{num}})$}
\newcommand{\mparentrule}{\top \Rightarrow y = f_{\top}(A'_{\textit{num}})}
\newcommand{\arule}{$p \Rightarrow y = f_p(A'_{\textit{num}})$}

\newcommand{\attcat}{$A_{\textit{cat}}$}
\newcommand{\mattcat}{A_{\textit{cat}}}

\newcommand{\mattnum}{A_{\textit{num}}}

\newcommand{\mattnump}{A'_{\textit{num}}}

\newcommand{\mcl}{\mathbf{cl}}
\newcommand{\mivd}{\mathit{iv}_{\mdataset}}

\AtBeginDocument{%
	\providecommand\BibTeX{{%
			\normalfont B\kern-0.5em{\scshape i\kern-0.25em b}\kern-0.8em\TeX}}}

\setcopyright{none}

\begin{document}
	
\title{\alg{}: Hierarchical Pattern-Aided Regression}


\author{Luis Galárraga}
\affiliation{%
	\institution{Inria}
	\country{France}
}
\email{luis.galarraga@inria.fr}

\author{Olivier Pelgrin}
\affiliation{%
	\institution{Aalborg University}
	\country{Denmark}}
\email{olivier@cs.aau.dk}

\author{Alexandre Termier}
\affiliation{%
	\institution{University of Rennes 1}
	\country{France}}
\email{alexandre.termier@irisa.fr}



\begin{abstract}
We introduce \alg{}, a novel pattern-aided regression method for tabular data containing both categorical and numerical attributes. \alg{} mines hybrid rules of the 
form $p \Rightarrow y = f(X)$ where $p$ is the characterization of a data 
region and $f(X)$ is a linear regression model on a variable of interest $y$. \alg{}
relies on pattern mining techniques to identify regions of the data
where the target variable can be accurately explained via local linear models. 
The novelty of the method lies in the combination of an enumerative approach to explore the space of regions and efficient heuristics that guide the search. 
Such a strategy provides more flexibility when selecting 
a small set of jointly accurate and human-readable hybrid rules that explain the entire dataset. 
As our experiments shows, \alg{} mines fewer rules than existing pattern-based regression methods while still attaining state-of-the-art prediction performance. 
\end{abstract}

\maketitle

\thispagestyle{empty}

\section{Introduction}
In the golden age of data, accurate numerical prediction models are of great utility in absolutely all disciplines.
The task of predicting a numerical variable of interest from the values of other variables --called features-- is known as regression analysis and the literature
is rich in this respect~\cite{piecewise-regression, cpxr, model-trees-with-splitting-nodes, regression-trees, boosting-trees, model-trees}. 
As data steadily complexifies, and the need 
for interpretable methods becomes compelling~\cite{survey-interpretability}, a line of research in regression analysis focuses 
on learning interpretable prediction models
on heterogenous data. By \emph{interpretable}, we mean models that provide a compact and comprehensible
explanation of the interaction between 
the features and the target variable, e.g., a linear function. 
By \emph{heterogeneous} data, we mean
data that can be hardly modeled by a single global regression function, but instead by a set 
of local models applicable to subsets of the data. The most prominent methods in this line are piecewise regression (PR, also called segmented regression)~\cite{piecewise-regression}, regression trees (RT)~\cite{regression-trees}, model trees (MT)~\cite{model-trees, model-trees-with-splitting-nodes} and contrast pattern-aided regression
(CPXR)~\cite{cpxr}. All these approaches mine \emph{hybrid rules} on tabular data such as the example in Table~\ref{table:example}. 
A hybrid rule is a statement of the form $p \Rightarrow y = f(X)$ where $p$ 
is a logical 
expression on categorical features such as $p : \mathit{property\text{-}type}=``\mathit{cottage}"$,
and $y = f(X)$ 
is a regression model for a numerical variable of interest, e.g., 
$\mathit{price} = \alpha + \beta \times \mathit{rooms} + \gamma \times \mathit{surface}$
that applies only to the region characterized by $p$, for instance, $\{x^1, x^2, x^3\}$ in Table~\ref{table:example}.

The advantage of methods based on hybrid rules is that they 
deliver statements that can be easily interpreted by humans. In contrast, they 
usually lag behind black-box methods such as gradient boosting trees~\cite{boosting-trees} or 
random forests~\cite{random-forests} in terms of prediction power. 
Indeed, our experiments show that existing pattern-aided regression methods have difficulties in providing satisfactory performance and interpretability simultaneously. 
On the one hand, methods such as RT or MT offer good prediction performance, but output many (long) rules: this makes them hard to read by a human user and thus less interpretable. On the other hand, CPXR outputs few simple rules (better interpretability), but its regression performance does not improve significantly over a simple global regression. 
The goal of this work is to reach a sweet spot where the produced set of hybrid rules is accurate and still simple enough to be grasped easily by a human user. 

\begin{figure*}
\begin{minipage}[t]{0.48\linewidth}
  \centering
    \includegraphics[width=\linewidth]{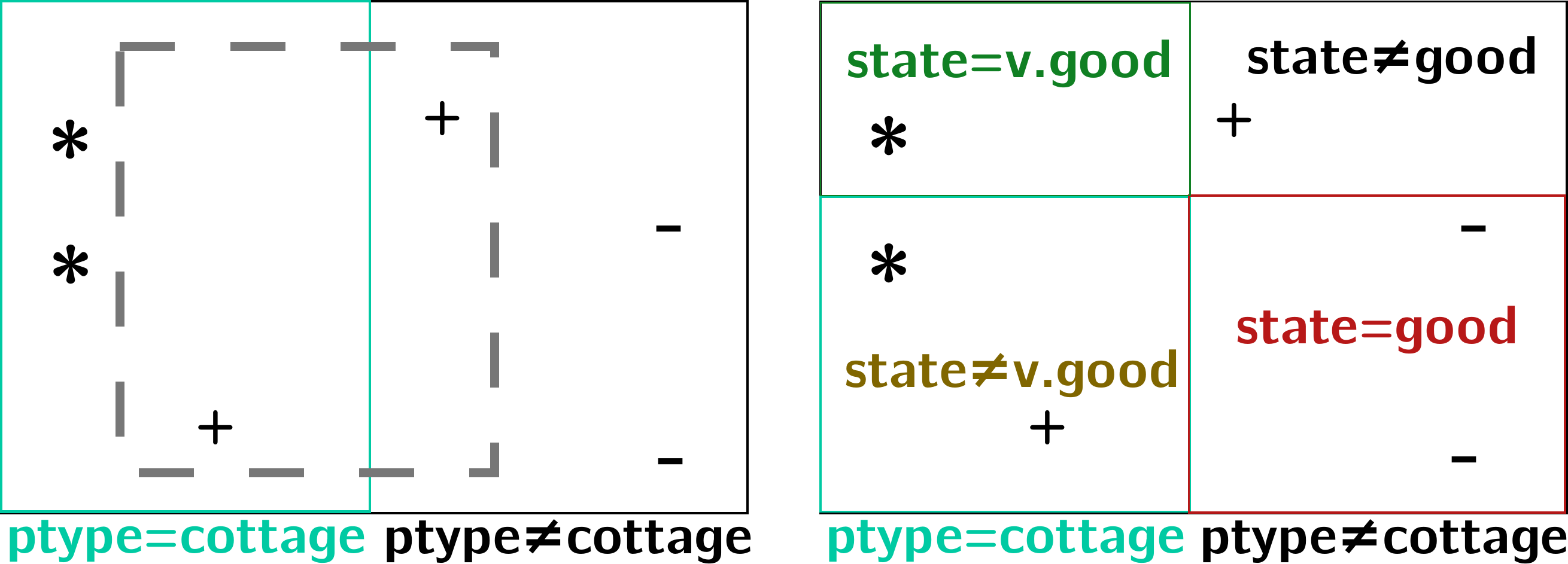}
    \captionof{figure}{Two steps of the exploration of regions by a regression tree learner induced on Table \ref{table:example}.}
    \label{fig:regression-tree-exploration}
\end{minipage}
\hspace{0.03\linewidth}
\begin{minipage}[t]{0.48\linewidth}
  \centering
    \includegraphics[width=\linewidth]{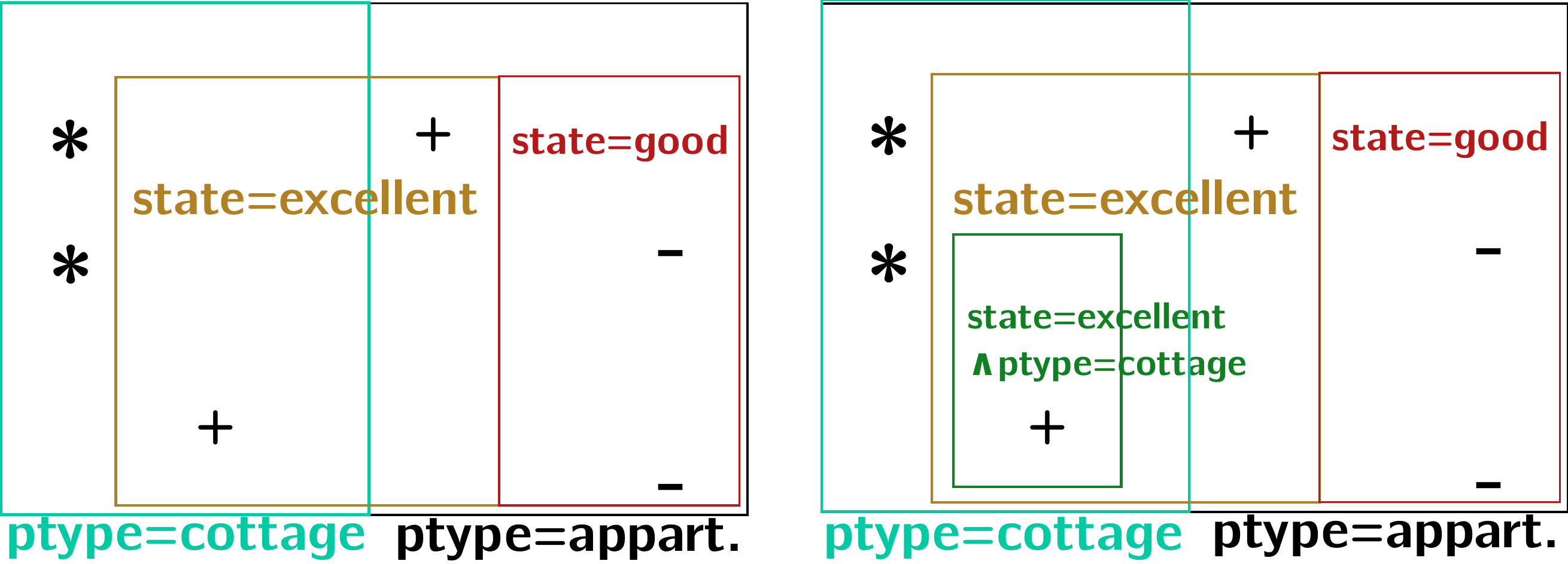}
    \captionof{figure}{Our hierarchical exploration applied to Table \ref{table:example}.}
    
    \label{fig:hipar-exploration}
\end{minipage}
\end{figure*}

Finding such a good set of hybrid rules is hard, because the search space of possible conditions (the left-hand side of the rules) is huge. 
Methods such as regression trees (RT) tackle this complexity with a greedy approach that refines rules with the best condition at a given stage. A simple regression tree for our example dataset is shown in Figure \ref{fig:regression-tree}, and the division of the data it entails is illustrated in Figure \ref{fig:regression-tree-exploration}. 
Regions with a high goodness of fit lying between two partitions, e.g., the dashed region on the left of Figure \ref{fig:regression-tree-exploration}, cannot be found even if they have short descriptions (here $state = ``\mathit{excellent}"$). They may only be described imperfectly by two longer patterns ($ptype=``\mathit{cottage}" \land state \neq ``\mathit{v.good}"$ and $ptype \neq ``\mathit{cottage}" \land state \neq ``\mathit{good}"$), which is less interpretable.
To avoid the shortcomings of a greedy exploration, CPXR~\cite{cpxr} proposes to enumerate the conditions of the rules using pattern mining techniques. More precisely, ~\cite{cpxr} 
applies {\em discriminative pattern mining} \cite{dong-discrim} to discover conditions describing subspaces in the data where a reference linear model yields the highest error, that is, data regions that may most likely benefit from local regression models. 
Due to its use of exhaustive enumeration, such approach can examine many alternative for rules and, unlike RT, authorizes overlap. 
A limitation of CPXR lies in its disregard of the data points where the error is not maximal but still high in absolute terms. Moreover, the rules found by the enumeration phase are then filtered by a greedy post-processing step.

Our main contribution lies in a novel strategy to explore the search space of hybrid rules. Such a strategy is hierarchical, as depicted in Figure~\ref{fig:hipar-exploration}, and is designed to find few short rules that fit the data. This gives rise to our method called \alg{}, which comprises two contributions:
\begin{itemize}[leftmargin=*]
    \item First, we design an hybrid rule enumeration algorithm that ouputs short and high-quality candidate rules. This algorithm is based on the enumeration structure of the state-of-the-art closed itemsets miner LCM~\cite{lcm}, which we augmented with several heuristics focused on the accuracy and compactness of the rules produced;
    \item Second, we frame the problem of selecting the best set of rules from any set of candidate rules as an Integer Linear Programming problem. 
    This allows for a modular and robust post-processing step to output a small set of high quality rules. 
\end{itemize}

Our experiments show that \alg{} reaches an interesting performance/intepretability compromise, providing as much error reduction as the best interpretable approaches but with one order magnitude fewer atomic elements (conditions) to examine for the analyst. 
Before detailing our approach and these experiments, we introduce relevant concepts and related work in the next two sections.

\begin{table}[t]
	\centering
	\caption{Toy example for the prediction of real estate prices based on the attributes of the property. The symbols
	*, +, - denote high, medium, and low prices respectively.}
	\begin{tabular}{>{\centering\arraybackslash}m{0.3cm}>{\centering\arraybackslash}m{2cm}>{\centering\arraybackslash}m{1.5cm}>{\centering\arraybackslash}m{0.8cm}>{\centering\arraybackslash}m{0.8cm}>{\centering\arraybackslash}m{1.2cm}}
	      \emph{id} 	& \emph{property-type} 	    	& \emph{state}   & \emph{rooms} & \emph{surface}	& \emph{price}	\\ 
		\toprule		
		{$x^1$} 	& {cottage}      		& very good	 &5         	& 120        		& 510k (*)              \\ 
		{$x^2$}	  	& cottage        		& very good    	 &3         	& 55        		& 410k (*)              \\
		{$x^3$}	  	& cottage     			& excellent 	 &3       	& 50          	   	& 350k (+)             \\ 
		{$x^4$} 	& apartment   			& excellent	 &5       	& 85          	   	& 320k (+)             \\ 
		{$x^5$} 	& apartment    			& good		 &4       	& 52         		& 140k (-)             \\
		{$x^6$} 	& apartment    			& good		 &3        	& 45        	       	& 125k (-)              \\ \bottomrule
	\end{tabular} 	
	\label{table:example}
\end{table}

%

\section{Preliminaries and Notation}
Pattern-aided regression methods assume tabular data with categorical and numerical attributes as in Table~\ref{table:example}. 
We define the notions of datasets, attributes, and patterns more formally in the following.
\subsection{Datasets} 
A dataset $D = \{ x^1, \dots, x^n \} \subseteq V^{|A|}$ is a set of $|A|$-dimensional points or observations, 
where $A$ is a finite set of attributes and each component of $x^i$ is associated to an attribute $a \in A$ with domain $V_a$. We denote the value of attribute $a$ for point $x^i$ by $x^i_a$. For instance, $x^{1}_{\mathit{state}} = ``\mathit{very\text{ }good}"$ in Table~\ref{table:example}. 
From a statistical perspective, attributes are random variables, thus in this work
the terms ``attribute'', ``feature'', and ``variable'' are used interchangeably.
A \emph{categorical} (also symbolic) attribute holds elements on which partial and total orders are meaningless. Examples are 
zip codes or property types as in Table~\ref{table:example}. 
A \emph{numerical} 
attribute, conversely, takes integer or real values and represents a measurable quantity 
such as a price or a temperature measure. Numerical attributes are the target of regression analysis.

\subsection{Patterns}
\label{subsec:patterns}
A pattern is a characterization of a dataset region (subset of points). An example is 
$p: \textit{property\text{-}type}=``\textit{cottage}" \land \textit{surface} \in (-\infty, 60]$ that 
describes the subset
$\{x^2, x^3\}$ in Table~\ref{table:example}. In this work we focus on conjunctive patterns
on non-negated conditions. These conditions take the form $a_i=v$ for categorical attributes 
or $a_j \in I$ for discretized numerical attributes, where $I$ is an interval
such as $(-\infty, \alpha)$, $[\alpha, \beta]$, or $(\beta, \infty)$, and $\alpha, \beta \in \mathbb{R}$. 

If $p$ is a pattern, we denote by $D_p$ its corresponding region on 
dataset $D$, and by $s_D(p) = |D_p|$ its support. We also define its relative support as $\bar{s}_D(p) = \frac{s_D(p)}{|D|}$. For instance, if $D$ is our example dataset from Table~\ref{table:example}, $s_D(p) = 2$ and $\bar{s}_D(p) = \frac{2}{6}$.
When the target dataset $D$ is implicit, we write $s(p)$ and $\bar{s}(p)$ for the sake of brevity.
A pattern $p$ is \emph{frequent} if $s(p) \ge \theta$, that is, if its associated region consists of at least $\theta$ data points for a given threshold $\theta$. 
A pattern is \emph{closed} if it is the maximal characterization of a region, i.e., no longer pattern can describe the same region. 
As each region can be described by a single closed pattern, we define the closure operator 
$\mathbf{cl}(p)$ of a pattern $p$ so that $\mathbf{cl}$ returns $D_p$'s associated
closed pattern. 
For instance, given the pattern $p : \mathit{state} =  ``\mathit{good}"$ 
characterizing the region $\{ x^5, x^6 \}$ in Table~\ref{table:example}, $\mathbf{cl}(p)$ is
$\mathit{state} =  ``\mathit{good}" \land \mathit{property\text{-}type}=``\mathit{apartment}"$,
because this is the maximal pattern that still describes $\{ x^5, x^6 \}$.
Given two subsets $D_1$ and $D_2$ of $D$ and a threshold $\gamma$, $p$ is a \emph{contrast} or \emph{emerging} pattern if 
(i) $\bar{s}_{D_1}(p) > 0$ and (ii) $\frac{\bar{s}_{D_1}(p)}{\bar{s}_{D_2}(p)} \ge \gamma$ or $\bar{s}_{D_2}(p)=0$, put differently, 
$p$ is a contrast pattern if it is at least $\gamma$ times (relatively) more frequent in $D_1$ than in $D_2$.

Last, we define the
interclass variance~\cite{traversing-lattices} of a pattern $p$ in $D$ w.r.t. a target numerical variable $y\in A$ 
as: \[ \mathit{iv}_D(p)= |D_p|(\mu_D(y) - \mu_{D_p}(y))^2 + |D_{\neg p}|(\mu_D(y) - \mu_{D_{\neg p}}(y))^2 \] 
\noindent In the formula $\mu_{\circ}(y)$ denotes the average of variable $y$ in a given dataset, whereas
$\neg p$ is the negation of pattern $p$ and $D_{\neg p} = D \setminus D_p$ is the complement of $D_p$. The interclass
variance is a measure of exceptionality.
A large $\mathit{iv}$ suggests that the values of $y$ 
in $D_p$ constitute a region of low variance that lies far from the variable's global mean, 
and is therefore a good candidate to learn local models.

\section{Related Work}
Having introduced a common notation, we now revisit the state-of-the-art in pattern-aided regression.
Furthermore, we discuss about two related paradigms for data analysis, namely subgroup discovery (SD) and exceptional model mining (EMM). 


\subsection{Piecewise Regression (PR)}
\cite{piecewise-regression} is among the first approaches for pattern-aided regression. PR splits the domain 
of one of the numerical variables, called the \emph{splitting variable},  
into segments such that the dataset 
regions defined by those segments exhibit a good linear fit for a \emph{target variable}. 
The splitting variable must be either ordinal\footnote{A special type of categorical attribute on which a total order on the values of its domain 
can be defined.} or numerical.
The regions are constructed via bottom-up 
hierarchical agglomerative clustering:   
Starting with clusters of size $\theta$, this bottom-up approach greedily picks the segment with the smallest 
residual average of squares, and fix it for the next iteration while \emph{declustering} the remaining points. 
Fixed clusters can be extended by incorporating adjacent points and other adjacent fixed clusters. The process stops when the number of isolated points drops below a threshold
or the goodness of fit does not improve with subsequent merging steps. 

Other variants of PR, such as~\cite{flirti}, focus on detecting regions of the space where the target variable correlates with polynomials of degree $n\neq1$ on the input features. That includes regions where we can predict a constant value for the target ($n=0$), or regions where polynomial regression is required ($n>1$).

PR usually outperforms single linear models on data with a multimodal distribution, 
however its limitations are manyfold. Firstly, it can only split the dataset based on one attribute at a time. 
Secondly, it cannot characterize data regions in terms of arbitrary categorical attributes. Thirdly, its greedy strategy does not guarantee
to find the best possible segmentation of the data~\cite{piecewise-regression}. 
PR models can be seen as sets of hybrid rules $z \in [v_1, v_2] \Rightarrow y = f(X)$, 
where the antecedent is an interval constraint on the splitting variable $z$.

\subsection{Tree-based Methods}
A regression tree~\cite{regression-trees} (RT) is a decision tree such that its leaves predict a numerical variable. 
Like decision trees,
RTs are constructed in a top-down fashion. At each step, the data is 
partitioned into two regions according to the condition that maximizes the intra-homogeneity 
of the resulting subsets w.r.t. the target variable (e.g., Figure~\ref{fig:regression-tree-exploration}). 
The conditions are defined on categorical and discretized numerical attributes.
This process is repeated while the subsets are large enough and its goodness of fit still improvable, otherwise
the learner creates a leaf that predicts the average of 
the target variable in the associated data region.
Model trees (MT)~\cite{model-trees, model-trees-with-splitting-nodes} associate linear functions to the leaves of the tree. 

\begin{figure}
	\centering
    \includegraphics[width=0.9\linewidth]{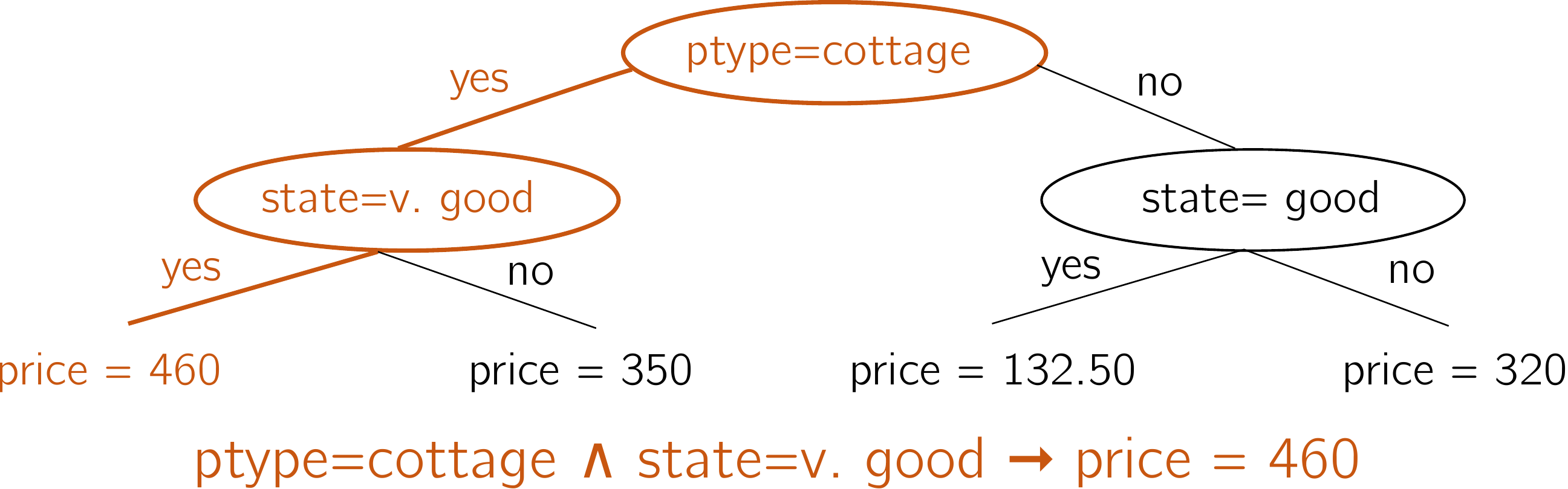}
    \caption{Regression tree learned to predict the price in Table~\ref{table:example}. Paths from the root to the leaves are hybrid rules.}
   \label{fig:regression-tree}
\end{figure}

We can mine hybrid rules from RT and MT 
if we enumerate every path from the root to a leaf (or regression node in~\cite{model-trees}) as depicted in Figure~\ref{fig:regression-tree}.
Unlike piecewise regression, RT and MT do exploit categorical features. Yet, their construction obeys 
a greedy principle: Data is split according to the criterion that maximizes the goodness of fit at a particular stage, 
and steps cannot be undone. This makes RT and MT prone to overfitting when not properly parameterized.  
More accurate methods such as random forests (RF) reduce this risk by learning tree ensembles that model the \emph{whole picture}. Alas, RF models are not interpretable. Some approaches~\cite{interpretable-rf, extracting-rules-from-rf} can extract representative rules from RF at the expense of accuracy. 

Our experiments in Section~\ref{sec:evaluation} confirm that 
RT and MT can make too many splitting steps yielding large and complex sets of rules,
even though we can attain a performance comparable to RF with fewer rules. 

\subsection{Contrast Pattern-Aided Regression}
\cite{cpxr} proposes CPXR, a method that mines hybrid rules on the regions 
of the input dataset where a global linear regression model performs poorly. 
First, CPXR splits the dataset
into two classes consisting of the data points where the global model yielded a 
large (LE) and a small error (SE) respectively. Based on this partitioning,
CPXR discretizes the numerical variables and mines contrast patterns~\cite{contrast-patterns} that characterize the difficult class LE. 
The algorithm then induces hybrid rules on the regions defined by those patterns.
After an iterative selection process, 
the approach reports a small set of hybrid rules with 
low overlap and good error reduction w.r.t. the global regression model. 
This set includes also a \emph{default model} 
induced on those points not covered by any rule.
Prediction for new cases in CPXR is performed by weighting the answers of all the rules that apply.  
The weights depend on the error reduction of the rules w.r.t. the global regression model. 

Despite its error-reduction-driven selection for rules, CPXR iterative search strategy is still greedy in nature.
Moreover, regions spanning across the classes LE and SE are disregarded, discretization is done once, and 
the search is restricted to the class of contrast patterns of the LE class 
(ignoring any error reduction in SE). While the latter decision keeps the 
search space under control, our experiments show that exploring the (larger) class of closed patterns allows 
for a significant gain in prediction accuracy with a reasonable penalty in runtime. 

\subsection{Related Paradigms} \label{subsec:sd-emm}
The problem of finding data regions with a high goodness of fit for a target variable is similar to the problem of subgroup discovery~\cite{sd} (SD). In its general formulation, 
SD reports subgroups --data regions in our jargon-- where the behavior of the target variable deviates notably 
from the norm. There exist plenty of SD approaches~\cite{sd-survey} tailored for different subgroup description languages, 
different types of variables and different notions of ``exceptionality''. For example,~\cite{sd} studies discretization techniques 
to deal with numerical attributes in subgroup descriptions, and shows the application of SD in diabetes diagnosis and fraud detection, i.e., to find the characterizations of subgroups with high incidence of diabetes and high fraud rate in mail order data. 

A more general framework, called Exceptional Model Mining (EMM)~\cite{emm,emm-cook-distance}, 
extends the notion of exceptionality to arbitrary sets of target variables. In this spirit EMM can find exceptional groups where the joint distribution of the target variables 
in a subgroup differs greatly from the global joint distribution. 
Finding exceptionally well-correlated subgroups in data can be framed as an SD or EMM task, nonetheless, these paradigms are concerned 
with reporting subgroups that are \emph{individually exceptional}. For this reason, EMM and SD methods are usually 
greedy and resort to strategies such as beam search to find a small set of exceptional subgroups. 
Conversely, we search for hybrid rules that are \emph{jointly exceptional}, that is, 
they (i) explain the whole dataset, and (ii) they jointly achieve good performance. 
While methods such as~\cite{sd} propose an average exceptionality score for sets of subgroups, 
such a simple score does not capture the requested synergy between sets of hybrid rules. 
Indeed, our experimental section shows that a SD-like selection strategy for hybrid rules yields 
lower performance gains than the strategy proposed in this paper.

\section{\alg{}}
\label{sec:alg}
In this section we describe our pattern-aided regression method called \alg{}, which is summarized in Algorithm~\ref{alg:hipar}. 
\alg{} mines hybrid rules of the form
$p \Rightarrow y = f_p(A'_{\mathit{num}})$ for a pattern $p$ characterizing a region $D_p \subseteq D$, and a
target variable $y$ on a dataset $D$ with attributes
$A = A_{\mathit{num}} \cup A_{\mathit{cat}}$. The sets 
$A_{\mathit{num}}$ and $A_{\mathit{cat}}$
define numerical and categorical attributes respectively, 
and $A'_{\mathit{num}} = A_{\mathit{num}} \setminus \{y\}$.  
Patterns and regions define a containment hierarchy that guides \alg{}'s search. This hierarchy is rooted at the empty pattern $\top$ that represents the entire dataset.
After learning a global linear model of the form \parentrule\; (also called the \emph{default model}), \alg{} operates in three stages: (i) initialization, (ii) candidates enumeration and (iii) rules selection. We elaborate on these phases in the following.


\begin{algorithm}
	\caption{\alg{}}
	\label{alg:hipar}
	\KwIn{a dataset: $\mathcal{D}$ with attributes $A_{\textit{cat}}$, $A_{\textit{num}}$ \\ 
		\hspace{25px} target variable: $y \in A_{\textit{num}}$ with $A'_{\textit{num}} = A_{\textit{num}} \setminus \{y\}$ \\ 
		\hspace{25px} minimum support threshold: $\theta$
	}
	\KwOut{a set $R$ of hybrid rules $p \Rightarrow y = f_p(A'_{\textit{num}})$}	
	Learn default hybrid rule $r_{\top} : \mparentrule$ from \dataset  \\
	$C := \textit{hipar-init}(\mdataset, y, \theta)$ \\
	$\mathcal{R} := \textit{hipar-candidates-enum}( \mathcal{D}, r_{\top}, C, \theta)$ \\    

	\Return $\textit{hipar-rules-selection}(\mathcal{R} \cup \{ r_{\top} \} )$
\end{algorithm}

\begin{algorithm}
	\caption{hipar-init}
	\label{alg:hipar-init}
	\KwIn{a dataset: \dataset~with attributes $A_{\textit{cat}}$, $A_{\textit{num}}$ \\ 
		\hspace{25px} target variable: $y \in \mattnum$ \\ 
		\hspace{25px} minimum support threshold: $\theta$
	}
	\KwOut{a set of frequent patterns of size 1}
	$C_{\textit{cat}} := \bigcup_{a \in \mattcat}{\{ c : a=v \;|\; s_D(c) \ge \theta  \}} $ \\
	$C_{\textit{num}} := \emptyset$ \\
	\For{$a \in \mattnum$}{
		$C_{\textit{num}} := C_{\textit{num}} \cup \{c : (a \in I) \in \textit{discr}(a, \mdataset, y)\;|\; s_D(c) \ge \theta \}$ \\
	}

	\Return $C_{\textit{cat}} \cup C_{\textit{num}}$
\end{algorithm}




\begin{algorithm}
	\caption{hipar-candidates-enum}
	\label{alg:hipar-candidates-enum}
	\KwIn{a dataset: \dataset~with attributes \attcat, $A_{\textit{num}}$  \\
		\hspace{25px} parent hybrid rule: $r_p: p \Rightarrow y = f_p(A'_{\textit{num}})$ \\
		\hspace{25px} patterns of size 1: $C$ \\
		\hspace{25px} minimum support threshold: $\theta$}
	\KwOut{a set $\mathcal{R}$ of candidate rules $p \Rightarrow y = f_p(A'_{\textit{num}})$}
	$\mathcal{R} := \emptyset$ \\
	$\mathcal{C'} := C$ \\
	$C_{n} := \{c \in C \;|\; c : a \in I\; \land a \in \mattnum \}$ \\
	$\nu := \text{k-th percentile of } \mivd \text{ in } C_{n}$ \\
	\For{$c' \in C'$}{
		$\hat{p} := p \land c'$  \\
		$C' := C' \setminus \{c'\}$ \\
		\If{$s_D(\hat{p}) \ge \theta \land \mivd(\hat{p}) > \nu $}{
			$p' = \mcl(\hat{p})$ \\
			$C' := C' \setminus p'$ \\
			\If{$p \text{ is the left-most parent of } p'$}{ 
				Learn $r_{p'} : p' \Rightarrow y = f_{p'}(\mattnump)$ on $\dataset_{p'}$  \\	
				\If{$m(r_{p'}) < m(r_{p^*})\; \forall p^* : p^* \;\text{is parent of}\; p'$}{
					$\mathcal{R} = \mathcal{R} \cup \{ r_{p'} \}$ \\
							
    				$C'_{n} := \emptyset$  \\ 
    				\For{$a \in \mattnump \setminus \textit{attrs}(p')$}{ 
    					$C'_{n} := \{c \in \textit{discr}(a, \mdataset_{p'}, y)\;|\; s_D(c) \ge \theta \} \cup C'_{n}$  \\
    				}
			
					$\mathcal{R} := \mathcal{R} \cup \textit{hipar-candidates-enum}(D, r_{p'}, (C' \setminus C_n) \cup C'_n, \theta)$ \\
    			}
			}			
		}
	}
	\Return $\mathcal{R}$
\end{algorithm}

\subsection{Initialization.} 
\label{subsec:initialization}
The initialization phase (line 2 in Algorithm~\ref{alg:hipar}) computes a set of frequent patterns that bootstrap \alg{}'s hierarchical search. 
We describe the initialization routine \emph{hipar-init} in Algorithm~\ref{alg:hipar-init}.
The procedure computes patterns of the form $a=v$ for categorical attributes (line 1), and $a \in I$ for numerical attributes (lines 2-4), where $I$ is an interval of the form $(-\infty, \alpha)$, $[\alpha, \beta]$, or $(\beta, \infty)$ (Section~\ref{subsec:patterns}). The intervals are calculated by discretizing the numerical attributes. 
The discretization is inspired on CPXR~\cite{cpxr}, that is, we first split the target variable into two classes, namely large value (LV) and small value (SV) and 
then run a routine~\cite{mdlp} that segments the domain of each numerical attribute so that the points 
in each segment are as pure 
as possible w.r.t. LV and SV. This way, \alg{} minimizes the variance of $y$ within the points that match a condition. The major difference with the discretization proposed in~\cite{cpxr} is that the classes LV and SV are defined w.r.t. the actual value of the target variable and not based on the residuals of a global regression model.
We remark that  \emph{hipar-init} enforces the bootstrapping patterns to be frequent, that is, their support must be higher than a user-defined threshold $\theta$ in the dataset (line 4 in Algorithm~\ref{alg:hipar-init})\footnote{When the intervals of an attribute $a$ are very imbalanced, that is, only one of the segments $c$ is large enough (e.g., $s(c) \ge \theta$ for a small $\theta$), it may be convenient to disregard the attribute for discretization.}. 

\subsection{Candidates Enumeration} \label{subsec:candidate-enumeration}
This stage
uses the patterns computed in the initialization step to explore the different regions of the dataset and learn local accurate candidate 
hybrid rules. These regions are characterized by closed patterns on categorical and discretized numerical variables. Our preference 
for closed patterns is based on two reasons. In the first place, and contrary to frequent and free patterns, closed patterns are not redundant. A region is characterized
by a unique closed pattern, whereas there may be a myriad of frequent or free patterns describing the exact same region. This property  prevents us from visiting the same region multiple times when traversing the search space. 
In the second place, closed patterns are expressive as they compile the maximal set of attributes that portray a region. This expressivity can be particularly useful in specialized domains when experts need to inspect the local regression models and identify \emph{all} the attribute values that correlate with the target variable. 

Inspired on methods for closed itemset mining~\cite{lcm}, the routine \emph{hipar-candidates-enumeration} in Algorithm~\ref{alg:hipar-candidates-enum} takes as input a hybrid rule \arule~learnt on a region characterized by $p$, and returns a set of hybrid rules defined on closed descendants of $p$. Those descendants are visited in a depth-first hierarchical manner as depicted in Figure~\ref{fig:searchtree}. Lines 1 and 2 in Algorithm~\ref{alg:hipar-candidates-enum} initialize the procedure by generating a working copy $C'$ of the set of conditions used to refine $p$ (loop in lines 5-18). At each iteration in line 5, Algorithm~\ref{alg:hipar-candidates-enum} extends the parent pattern $p$ with a condition $c' \in C'$ and removes $c'$ from the working copy (line 7). After this refinement step, the routine proceeds in multiple phases detailed below.

\subsubsection{Pruning}
In line 8, Algorithm~\ref{alg:hipar-candidates-enum} enforces thresholds on support and interclass variance for the newly refined pattern $\hat{p} = p \land c'$. 
The support threshold $\theta$ serves two purposes. 
First, it prevents us from learning rules on extremely small subsets and incurring overfitting -- a problem frequently observed in unbounded regression trees. Second, since \alg{}'s search space is exponential in the number of conditions, a threshold on support allows us for pruning, and hence lower runtime. 
While a reasonable support threshold 
can mitigate the pattern explosion in low-dimensional datasets, the on-the-fly discretization of the numerical attributes carried out in lines 16-17 may contribute 
with a large number of new frequent conditions. 
On these grounds \alg{} applies a second level of
pruning by means of a threshold $\nu$ on the interclass variance $\mathit{iv}$ as proposed in~\cite{traversing-lattices}. We highlight the heuristic nature 
of using $\mathit{iv}$ for pruning, since this metric lacks the anti-monotonicity of support. 
That means that a region with an $\mathit{iv}$ below the threshold
can contain sub-regions with high $\mathit{iv}$ that will not be explored. 
Having said that, thresholding
on interclass variance proves effective at keeping the size of the search space under control
with no impact on prediction performance. We set $\nu$ empirically to the 85-th percentile of the interclass variance 
of the patterns derived from the discretization of the numerical features (lines 3 and 4). 
Lower percentiles did not result in better performance in our experimental datasets.

\subsubsection{Closure Computation}
If a refinement $\hat{p} = p \land c'$ passes the test in line 8, \alg{} computes its corresponding closed pattern $p'$ in line 9. Since the closure operator may add further conditions besides $c'$, line 10 ensures that those conditions are not considered for future refinements\footnote{Our abuse of notation treats $p$ as a set of conditions}. Next, the check in line 11 guarantees that no path in the search space is explored more than once. This is achieved by verifying whether pattern $p$ is the leftmost parent of $p'$. In Figure~\ref{fig:hipar-exploration}, this check ensures that the sub-tree rooted at the node $\textit{ptype}=``\textit{cottage}" \land \textit{ptype}=``excellent"$ is explored only once, in this case from its leftmost parent $\textit{ptype}=``\textit{cottage}"$.

 \begin{figure}
    \includegraphics[width=\linewidth]{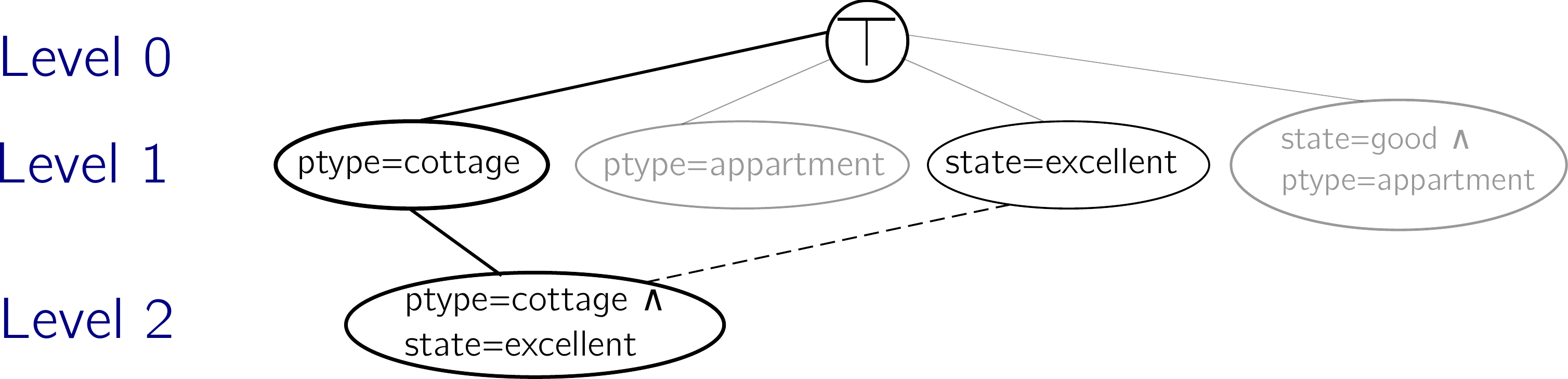}
    \caption{\alg{} hierarchical region exploration tree.}
        \label{fig:searchtree}
\end{figure}
 
\subsubsection{Learning a Regression Model} 
In line 12, Algorithm~\ref{alg:hipar-candidates-enum} learns a hybrid rule $p' \Rightarrow y = f_{p'}(A'_{\mathit{num}})$ from the data points that match $p'$. Before being accepted as a candidate (line 14), this new rule must pass a test in the spirit of Occam's razor (line 13): among multiple hypotheses of the same prediction power, the simplest should be preferred. This means that if a hybrid rule defined on 
 region $D_{p'}$ does not predict better than the hybrid rules defined in the super-regions of $D_{p'}$, 
 then the rule is redundant, because we can obtain good predictions with more general and simpler models. 
 In this line of thought, \alg{} adds the newly created hybrid rule as a candidate if 
 it performs better than the hybrid rules induced on the immediate ancestors of $p'$ in the hierarchy. 
 Performance is defined in terms of an error metric $m$ (e.g., RMSE).
 This requirement makes our search diverge from a pure DFS as 
 shown in Figure~\ref{fig:searchtree}. 
 For instance, the performance 
 test for the node \emph{ptype}=``\emph{cottage}'' $\land$ \emph{state}=``\emph{excellent}'' requires us 
 to visit its parent \emph{state}=``\emph{excellent}'' earlier than in a standard DFS. 
 
\paragraph{DFS Exploration.} 
The final stage of the routine \emph{hipar-candidates-enum} discretizes the numerical variables not yet discretized in $p'$ (lines 16-17)\footnote{\emph{attrs}(p) returns the set of attributes present in a pattern} and uses those conditions to explore the descendants of $p'$ recursively (line 18). We remark that this recursive step could be carried out regardless of whether $p'$ passed or not the test in line 13: Error metrics are generally not anti-monotonic, thus the region of a rejected candidate may still contain sub-regions that yield more accurate hybrid rules. Those regions, however, are more numerous, have a lower support, and are characterized by longer patterns. Given \alg{}'s double objective for accuracy and interpretability, recursive steps become less appealing as we descend in the hierarchy. This early stopping heuristic had no impact on prediction accuracy in our experiments.  

\subsection{Rule Selection} 
\label{subsec:ruleSelection}
The discovered set $\mathcal{R}$ of candidate rules $r_p : p \Rightarrow y = f_p(A'_{\mathit{num}})$, being generated from a combinatorial enumeration process, is likely to be too large for presentation to a human user.
Thus, \alg{} carries out a selection process (line 4 in Algorithm~\ref{alg:hipar}) that picks a
subset (of rules) of minimal size and 
minimal joint error 
such that as many observations as possible in \dataset~are covered. 
We formulate these multi-objective desiderata as an integer linear program (ILP): 

\begin{equation}\label{eq:formulation}
\begin{aligned}
& \text{min} \; \sum_{r_p \in \mathcal{R}}{-\alpha_p \cdot z_p} + \; \sum_{r_p, r_q \in \mathcal{R}, p \neq q}{(\omega \cdot \mathcal{J}(p, q) \cdot (\alpha_p + \alpha_q))\cdot z_{pq}} \;\;\; \\
& \text{s.t.} \;\;\; \;\;\; \sum_{r_p \in \mathcal{R}} z_p \ge 1 \;\;\;\\
& \;\;\; \;\;\; \;\;\; \; \forall r_p, r_q \in \mathcal{R}, p \neq q : z_p + z_q - 2z_{pq} \le 1  \;\;\;  \\
& \;\;\; \;\;\; \;\;\; \; \forall r_p, r_q \in \mathcal{R}, p \neq q : z_p, z_{pq} \in \{0, 1\} \;\;\;
\end{aligned}
\end{equation} 
\noindent Each single rule $r_p \in \mathcal{R}$
is associated to a variable $z_p$ that takes either 1 or 0 depending on whether 
the rule is selected or not. The first constraint guarantees a non-empty set of rules. The term $\alpha_p =\bar{s}(p)^\sigma \times \bar{e}(r_p)^{-1}$ 
is the support-to-error trade-off of the rule. Its terms $\bar{e}(r_p) \in [0,1]$ and $\bar{s}(p) \in [0, 1]$ correspond to 
the normalized error and 
normalized support of rule $r$ calculated as follows:

\begin{minipage}{0.45\linewidth} 
\begin{equation} 
\label{eq:normalized-error} \bar{e}(r_p) = \frac{m(r_{p})}{\sum_{r_{p'} \in \mathcal{R}}{m(r_{p'})}}
\end{equation} 
\end{minipage} 
\begin{minipage}{0.45\linewidth} 
\begin{equation} 
\bar{s}(p) = \frac{s(p)}{\sum_{r_{p'} \in \mathcal{R}}{s(p')}}
\end{equation} 
\end{minipage} \vspace{0.5cm}

\noindent In plain English, $\bar{e}(r_p)$ is the error of rule $r_p$ according to an error metric $m$ divided by the sum of 
the errors of all rules in the set of candidates. The normalized support $\bar{s}(p)$ is calculated in the same spirit.
%
It follows that the objective function rewards rules with small error defined 
on regions of large support. This latter property accounts
for our maximal coverage desideratum and protects us from overfitting. 
The support bias $\sigma \in \mathbb{R}_{\ge 0}$ is a meta-parameter that controls the importance of 
support in rule selection. 
When $\sigma=0$ the solver disregards support, whereas larger values define
different trade-offs between accuracy and support.
Due to our hierarchical exploration, the second term 
in the objective function penalizes sets with rules defined
on overlapping regions.
The overlap is measured via the Jaccard coefficient on the regions of each pair of rules, i.e., $$\mathcal{J}(p, q) = \frac{|D_p \cap D_q|}{|D_{p} \cup D_{q}|}.$$

\noindent If two rules $r_p$, $r_q$ are selected, i.e., $z_p$ and $z_q$ are set to 1, the second family of constraints enforces the variable
$z_{pq}$ to be 1 and pay a penalty proportional to $\omega \times (\alpha_p + \alpha_q) $ times the degree of overlap between $D_p$ and $D_q$ in
the objective function. The overlap bias $\omega \in \mathbb{R}_{\ge 0}$ controls the magnitude of the penalty.
Values closer to 0 will make the solver tolerate overlaps and choose more rules.  
The solution to Equation~\ref{eq:formulation} is a set of accurate hybrid rules 
$R \subseteq \mathcal{R}$ that can be used as a prediction model for the target variable $y$.

\subsection{Prediction with \alg{}} \label{subsec:prediction}
To use the rules reported by \alg{} as a prediction model, we have to define a procedure to deal with 
overlapping rules as well as with orphan data points. 
A rule $r : p \Rightarrow y = f_p(A'_{\mathit{num}})$ \emph{is relevant to} or \emph{covers} a seen or unseen data point
$\hat{x}$ if the condition defined by the pattern $p$ evaluates to true on $\hat{x}$.
If $\hat{x}$ is not covered by any hybrid rule, \alg{} uses the default regression model $r_{\top}$ 
to produce a prediction. Otherwise, \alg{} returns a weighted sum of the predictions of 
all relevant rules of $\hat{x}$ (as done in~\cite{cpxr}). The weight $\alpha_{p, \hat{x}}$ associated to a rule 
$r_p$ when predicting $\hat{x}$ is calculated as:
\[
\alpha_{p, \hat{x}} = \frac{\bar{e}(r_p)^{-1}}{\sum_{r_{p'} \in \Phi(\hat{x})}{\bar{e}(r_{p'})^{-1}}}
\]
$\Phi(\hat{x})$ denotes the set of rules that cover $\hat{x}$, and $\bar{e}(r_p)$ is the rule's normalized error in the training set 
(Equation~\ref{eq:normalized-error}).

\section{Evaluation}
\label{sec:evaluation}
We evaluate \alg{} on the dimensions of prediction accuracy, interpretability, and runtime through three rounds of experiments. In the first round (Section~\ref{subsec:impact-parameters}), we measure the impact
of \alg{}'s parameters on our evaluation aspects. The second round compares \alg{} with state-of-the-art regression methods (Section~\ref{subsec:comparison-state-of-the-art}). In a third round, we carry out an anecdotal evaluation by showing and analyzing some of the rules mined by \alg{} on well-studied use cases (Section~\ref{subsec:anecdotal_evaluation}). Section~\ref{subsec:experimental-setup} provides a preamble by describing our experimental setup.

\subsection{Experimental Setup}
\label{subsec:experimental-setup}

\subsubsection{\alg{}'s implementation.} \label{subsec:implementation} We implemented \alg{} in Python 3 with scikit-learn\footnote{\url{http://scikit-learn.org}}. In addition to the parameters described in Algorithm~\ref{alg:hipar}, our implementation accepts as input a support bias $\sigma$,
an overlap bias $\omega$ (with default values $\sigma=1$ and $\omega=1$), an error metric $m$, and a type of regression model. We shed light on how to tune $\sigma$ and $\omega$ in Section~\ref{subsec:impact-parameters}. We evaluate \alg{} with two error metrics, namely the root mean square error (RMSE) and the median absolute error (MeAE). Moreover, we test \alg{} with two methods for sparse linear regression, namely 
OMP~\cite{omp} and LASSO~\cite{lasso}. Sparse linear models optimize a regularized objective function that instructs the regressor to use 
as few non-zero coefficients as possible.
This choice makes linear functions more legible and conforms to our interpretability requirement. 
Since there is no clear winner between OMP and LASSO, we configured \alg{} to learn, for each pattern $p$, hybrid rules with both methods and keep 
the rule with the lowest error in a test set of 20\% the size of $D_p$. In regards to the discretization of the numerical attributes (Section~\ref{subsec:initialization}), 
\alg{} uses the MLDP algorithm~\cite{mdlp}. This method resorts to the principle of minimum description length (MDL) to obtain simple multi-interval discretizations of the numerical variables. 
The source code of \alg{} is available at \url{http://gitlab.inria.fr/lgalarra/hipar}. 

\subsubsection{Opponents.} \label{subsec:opponents}
We compare \alg{} to multiple regression methods comprising: 
\begin{itemize}[leftmargin=*]
 \item Three pattern-aided regression methods, namely CPXR~\cite{cpxr}, regression trees (RT)~\cite{regression-trees}, 
 and model trees (MT)~\cite{model-trees}.
 \item Three accurate black-box methods: random forests (RF)~\cite{random-forests}, gradient boosting trees (GBT)~\cite{boosting-trees}, 
 and rule fit (Rfit)~\cite{rulefit}.
 \item \alg{} when all rules output by the enumeration stage are selected ($\omega=0.0$, called $\alg_f$), and 
 \alg{} with a rule selection in the spirit of subgroup discovery: the top $q$ rules with the best 
 support-to-error trade-off are reported. 
 The parameter $q$ is set to the average number of rules output by \alg{} in cross-validation. We denote
 this opponent by $\alg_{sd}$. 
 \item Two hybrid methods resulting from the combination of \alg{}'s enumeration phase 
 with Rfit, and Rfit's rule generation with \alg{}'s rule selection. We denote these methods by
 \alg{}+Rfit, and Rfit+\alg{} respectively. These competitors are designed to evaluate the two phases of \alg{} in isolation. 
\end{itemize}

\noindent The opponents that can work with any linear regression method, namely, MT, CPXR, and $\alg_{sd}$, are reported
with the best performing method (either LASSO or OMP).
Since there is no available implementation
of CPXR, we implemented the algorithm in Python 3 with scikit-learn (the implementation is provided with \alg{}'s source code). We use the scikit-learn implementation of RT, whereas for MT we use the implementation available at \url{https://is.gd/Gk9Y20} (based on CART).
By default, RT and MT do not impose constraints on the size of trees. Hence, they can yield large numbers of complex hybrid rules that are hardly interpretable.
For this reason we test compact variants of RT and MT obtained by allowing at most $q+1$ leaves in the trees. 
We set $q$ equals the number of rules found by \alg{} in cross-validation. 
This way, we can compare the accuracy of \alg{} and tree-based methods at a similar level of interpretability. We denote
these two settings by RT$_H$ and MT$_H$ respectively.
For GBT and RF we use the implementations available in scikit-learn, whereas for Rfit we use
the source code provided by the authors\footnote{\url{https://github.com/christophM/rulefit}}. 
All black-box methods are ensemble methods on regression trees, that is, 
they rely on the answers of multiple trees (called \emph{estimators}) to compute a final prediction.
The main hyper-parameters -- that are not fixed by the requirements of the experimental setup -- are tuned using \emph{hyperopt}\footnote{\url{https://github.com/hyperopt/hyperopt}} for all competitors. That includes, for instance, the support threshold for CPXR and RT, or the maximal tree depth for RT, RF, and Rfit.
All the experiments were run on a computer with a CPU Intel Core i7-6600U (@2.60GHz), 16GB of RAM, and Fedora 
26 as operating system.

\subsubsection{Datasets.}
We test \alg{} and the competitors in 7 out of the 50 datasets used to evaluate CPXR~\cite{cpxr}. 
Neither the authors of CPXR nor the original collectors of the datasets~\cite{iie} could provide us with the data, thus
we collected by ourselves the 7 datasets that are publicly available at the UCI repository\footnote{\url{http://archive.ics.uci.edu/ml/index.php}}:
\emph{abalone}, \emph{cpu}, \emph{houses}, \emph{mpg2001}, \emph{servo}, \emph{strikes}, and \emph{yatch}. 
We also downloaded 8 additional datasets from Kaggle\footnote{\url{http://kaggle.com}}, namely \emph{cb\_nanotubes}, \emph{fuel\_consumption}, 
\emph{healthcare} (we used a sample due to its large size), \emph{optical}, \emph{wine}, \emph{concrete}, \emph{beer\_consumption} and 
\emph{admission}. 
These datasets match the keyword ``regression'' in the Kaggle's search engine (as of 2019), and 
define meaningful regression tasks.  
In addition, our anecdotal evaluation in Section~\ref{subsec:anecdotal_evaluation} relies on the results presented in~\cite{emm-cook-distance} on the datasets \emph{giffen} and \emph{wine2}. 
Table~\ref{table:datasets} provides details about the different datasets.
The datasets and their descriptions are available for download with \alg{}'s source code. 

 \begin{table}
 	\centering
 	\caption{Experimental datasets.}
 	\begin{tabular}{>{\centering\arraybackslash}m{2.5cm}>{\centering\arraybackslash}m{1.2cm}>{\centering\arraybackslash}p{1.6cm}>{\centering\arraybackslash}m{1.8cm}}
 		\emph{Dataset} 		& \emph{\# obs.} 	& \emph{\# cat. attrs} & \emph{\# num. attrs} \\ 
 		\toprule		
 		abalone 		   	& 4177		        & 1					& 8     \\ 
 		admission		   	& 500				& 2					& 7	\\
 		beer\_consumption	& 365				& 2					& 5	\\
 		cb\_nanotubes		& 10722				& 0					& 8 	\\
 		concrete		   	& 1030				& 0					& 9	\\	
 		cpu 			   	& 209		        & 2		 			& 5                   \\
 		fuel\_consumption	& 389				& 5					& 5	\\
 		giffen				& 6668				& 6					& 47	\\		
 		houses 			   	& 6880	        	& 0		 			& 9                 \\ 
 		mpg2001 	        & 852		        & 7					& 10                  \\ 
 		servo 			   	& 167		        & 2					& 3                 \\
 		strikes 		   	& 625		        & 1					& 6                 \\ 		
 		healthcare	   	   	& 518				& 6					& 192	\\
 		optical   	  	   	& 641				& 3					& 10	\\ 
 		wine			   	& 6498	    		& 1					& 12     \\
 		wine2			   	& 9600	    		& 6					& 5     \\ 		
 		yatch 		   	   	& 308		        & 0					& 7                \\ \bottomrule
 	\end{tabular} 	
 	\label{table:datasets}
 \end{table}

\subsubsection{Metrics.} We evaluate the prediction accuracy of the 
different approaches in terms of the root mean square error (RMSE) and the median absolute error (MeAE). 
We report the error reduction w.r.t. a baseline  
non-regularized linear model $\mathcal{B}$ on the entire dataset. The reduction $\rho$
of a regression model $\mathcal{M}$ for an error metric $m$ 
is calculated as follows:
\[
\rho = \frac{m(\mathcal{B}) - m(\mathcal{M})}{m(\mathcal{B})} \times 100
\]
Since our goal is to mine sets of human-readable rules, we also evaluate \alg{}'s rules in terms of interpretability. 
We remark, however, that this notion is subjective and may depend on factors such as the 
user's background. 
Nevertheless, it is widely accepted that analyzing 100
rules with 20 conditions each is more challenging than grasping the information in 5 rules with 3 conditions each. 
In this line of thought, 
and due to the diversity of domains of our datasets, we use complexity as a proxy for interpretability. Hence, we conduct a quantitative analysis based on the
number of ``elements'' in a model.
An element is either a condition on a categorical attribute or 
a numerical variable with a non-zero coefficient in a regression function. For tree-based models, we count each non-leaf node as an element, whereas a leaf contributes with multiple elements:
one per variable present in the associated regression function. For example, the regression tree in Figure~\ref{fig:regression-tree} consists of 7 elements. 

\subsection{Impact of Parameters}
\label{subsec:impact-parameters}

\subsubsection{Support Threshold.} \label{subsec:impact-of-support} 
\begin{figure}
	\centering
	\includegraphics[width=0.95\linewidth]{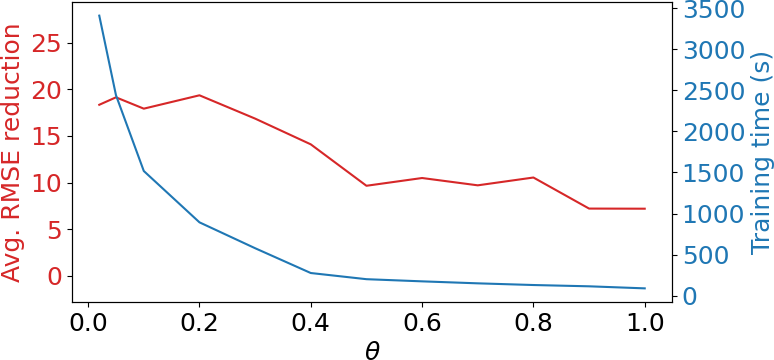}
	\caption{Support threshold vs. error reduction and training time in seconds.}
	\label{fig:support_impact}
\end{figure}

\begin{figure}
	\centering
	\includegraphics[width=0.90\linewidth]{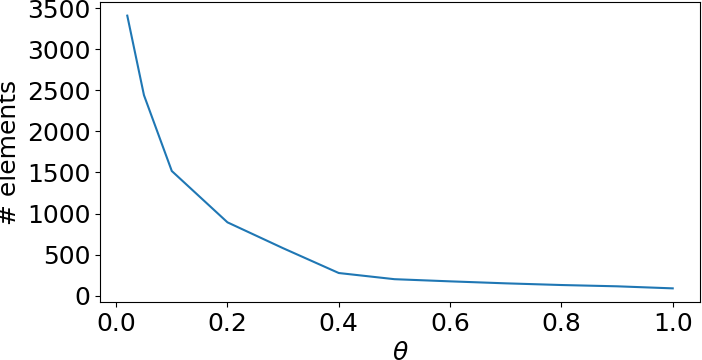}
	\caption{Support threshold vs. average \# of elements.}
	\label{fig:support_impact2}
\end{figure}

The minimum support threshold $\theta$ controls the exhaustivity of \alg{}'s candidates enumeration (Alg.~\ref{alg:hipar-candidates-enum}). Lower values make \alg{} report more rules defined on very specific regions. Thus, $\theta$ has a direct impact on \alg{}'s runtime and complexity as depicted in Figures~\ref{fig:support_impact} and~\ref{fig:support_impact2} where we plot relative support vs. \alg{}'s average RMSE reduction, average training runtime, and average number of elements of a round of (10-fold) cross-validation across the experimental datasets. We observe that values between 0.1 and 0.3 offer a good trade-off between prediction accuracy, runtime, and interpretability. Support thresholds below 0.1 increase prediction accuracy marginally at the price of doubling runtime and model complexity. Conversely, as $\theta$ approaches 1, \alg{} tends to select only the default rule $r_{\top}$ becoming tantamount to a regularized linear regression.

\subsubsection{Support and Overlap Biases.} 
\begin{figure}
	\centering
	\includegraphics[width=0.92\linewidth]{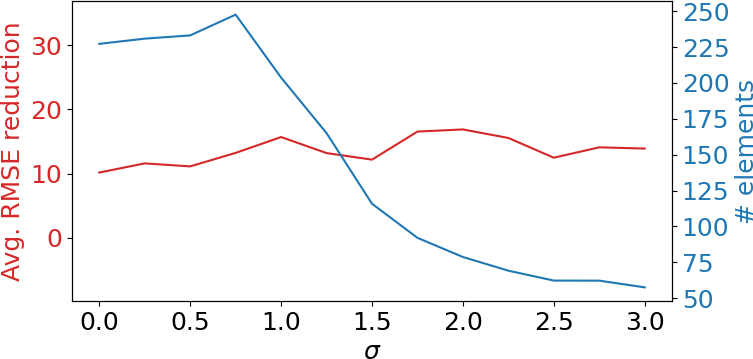}
	\caption{Support bias vs. error reduction and \# of elements.}
	\label{fig:evolution_sigma}
\end{figure}

\begin{figure}
	\centering
	\includegraphics[width=\linewidth]{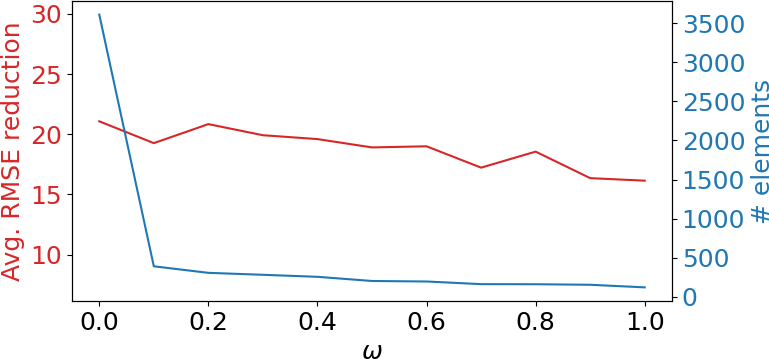}
	\caption{Overlap bias vs. error reduction and \# of elements.}    
	\label{fig:evolution_omega}
\end{figure}

\begin{figure}
	\centering
	\includegraphics[width=0.90\linewidth]{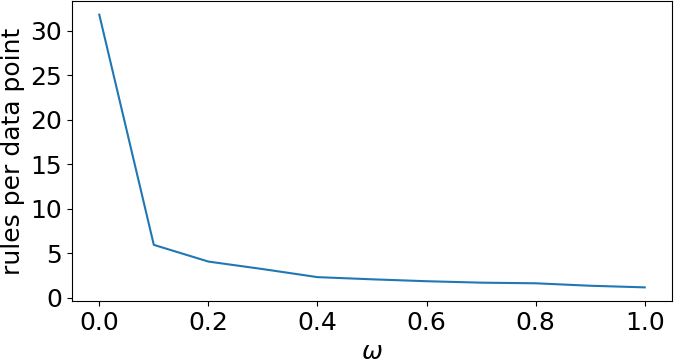}
	\caption{Overlap bias vs. average \# of rules per point.}    
	\label{fig:evolution_omega2}
\end{figure}

Figures~\ref{fig:evolution_sigma} and~\ref{fig:evolution_omega}
show the impact on \alg{}'s RMSE reduction and  number of elements of the arguments that govern the rule selection, 
that is, the support and overlap biases $\sigma$ and $\omega$ (Section~\ref{subsec:ruleSelection}). We plot the averages across our 
experimental datasets when fixing one parameter and varying the other one. We set $\theta=0.02$ in order to guarantee a large number of candidates as input to the selection phase. 
We observe that the RMSE reduction is mostly insensitive to changes in $\sigma$ and -- to a lesser extent -- $\omega$. 
Contrary to the error reduction, the number of elements always tends to decrease as the parameters take higher values. 
This corroborates our intuition, so to say, that many of the rules induced in the exploration phase are not essential for accurate prediction.
Small values of $\sigma$ downplay the role of support in the importance of rules since $\bar{s}(p)^{\sigma} \rightarrow 1$ as $\sigma \rightarrow 0$ (Section \ref{subsec:ruleSelection}). This rewards rules with low error regardless of their coverage. For low support thresholds, a small $\sigma$ translates into large sets of highly specific rules
that are unlikely to overlap. As $\sigma$ increases, specific rules are penalized and their selection becomes less likely (recall that $0 < \bar{s}(p) \le 1$).
On the other hand, a large $\omega$ means that sets of overlapping rules 
are highly penalized, and points are covered by fewer rules as depicted in Figure~\ref{fig:evolution_omega2}. 
This makes \alg{} closer to compact RTs and MTs as 
it forces rules to cover disjoint regions. Based on our observations, we recommend to set $\omega=1$ and $\sigma \ge 1$ depending on the need for more general or more specific rules.


\subsection{Comparison with the State of the Art.}
\label{subsec:comparison-state-of-the-art}
\subsubsection{Accuracy and Complexity Evaluation}
\label{subsubsec:comparison-accuracy-and-interpretability}
\begin{figure}
  \centering
    \includegraphics[width=\linewidth]{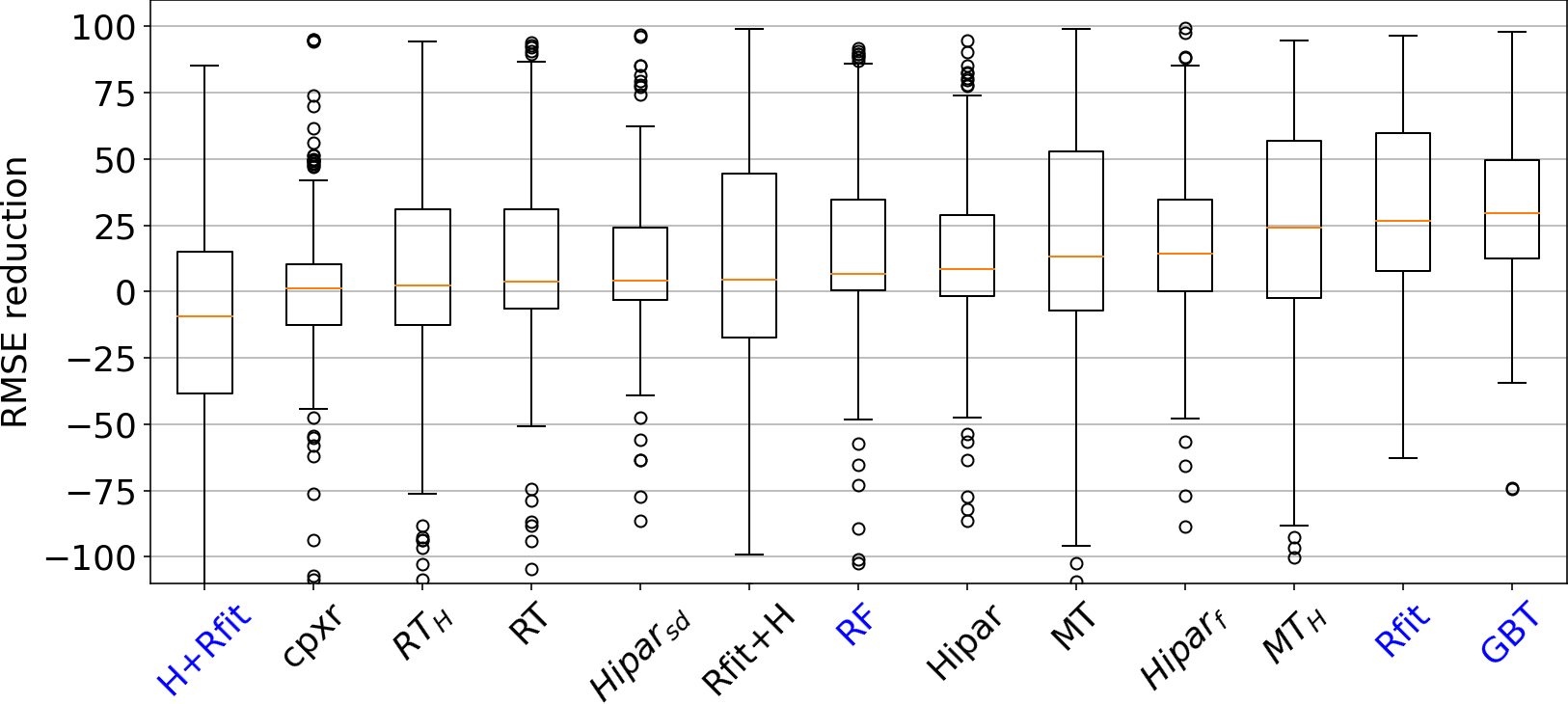}
    \caption{Mean RMSE reduction in cross-validation. Black-box methods are in blue}
    \label{fig:whiskers_rmse}
\end{figure}

\begin{figure}
  \centering
    \includegraphics[width=\linewidth]{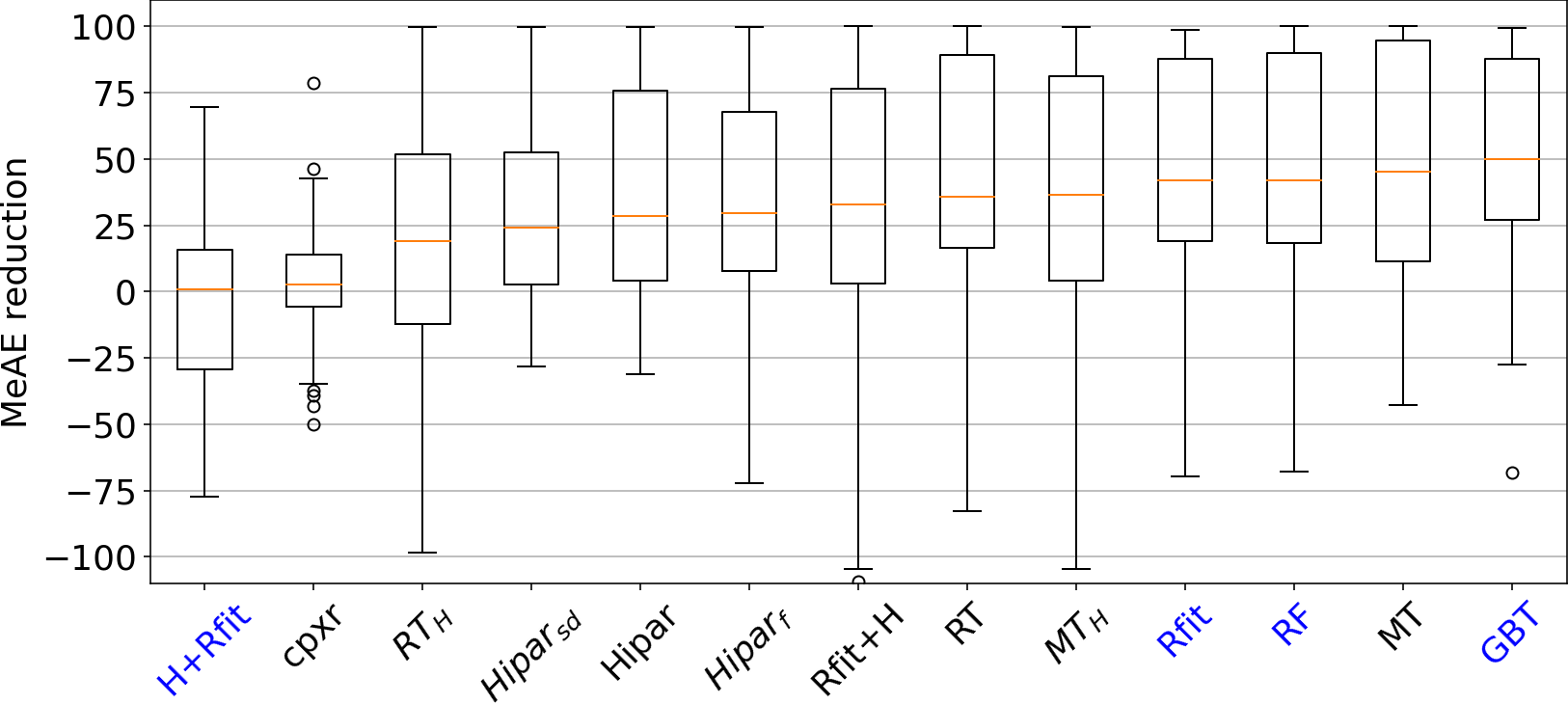}
    \caption{Median MeAE reduction in cross-validation.}
    \label{fig:whiskers_mae}
\end{figure}

Figures~\ref{fig:whiskers_rmse} and~\ref{fig:whiskers_mae} depict the mean RMSE and median MeAE reductions in 
10-fold cross-validation for the different methods on our experimental datasets. The methods are sorted
by the median reduction of all executions. 

\begin{figure}
  \centering
  \includegraphics[width=\linewidth]{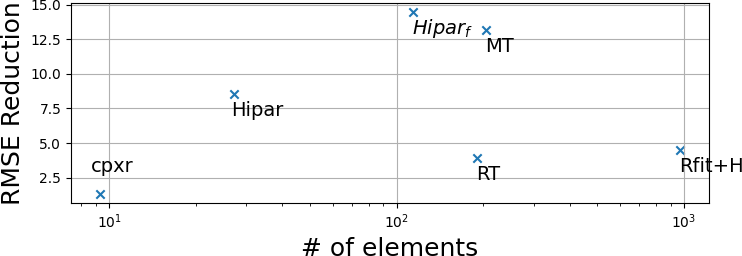}
  \caption{Trade-off between number of elements and RMSE reduction of the different methods.}
  \label{fig:tradeoff}
\end{figure}

We first note that the black-box methods, i.e., GBT, Rfit, and
RF (in blue) rank higher than the interpretable methods \alg{}, MT or RT in terms of error reduction. 
The unbounded tree-based approaches usually achieve good performance, however this comes at the expense of complex sets of rules as depicted in Figure~\ref{fig:tradeoff} for the RMSE (the MeAE exhibits the same behavior). 
If we tune the maximum number of leaves in the trees using \alg{} -- denoted by RT$_H$ and MT$_H$ --, we observe a positive impact on the RMSE for MT, whereas RT see a slight drop in performance (Figure~\ref{fig:whiskers_rmse}).  
This suggests that the greedy exploration of RT and MT may lead
to unnecessary splitting steps where good performance is actually attainable with fewer rules. Conversely, setting a limit in the number of leaves has a deleterious effect on the MeAE.

We observe that \alg{}'s median RMSE reduction is comparable to unbounded MT and $\alg_f$. Yet, \alg{} outputs one order of magnitude fewer
elements as shown in Figure~\ref{fig:tradeoff}, thanks to our rule selection step. Besides, \alg{}'s behavior is  
more stable than the tree-based methods, i.e., it yields fewer extreme values, and is comparable to RF (Figure~\ref{fig:whiskers_rmse}).   
The situation is slightly different for the MeAE (Figure~\ref{fig:whiskers_mae}), where \alg{} has a lower median reduction than MT and competes with Rfit+\alg{}, although the latter two methods exhibit a larger variance. This shows that \alg{}'s rule selection can work with other rule generation methods -- tree ensembles as implemented by Rfit. The greedy rule selection implemented by $\alg_{\textit{sd}}$ yields poorer results than standard \alg{} and $\alg_f$.  

We also observe that CPXR and \alg{}+Rfit lie 
at the bottom of the ranking in Figures~\ref{fig:whiskers_rmse} and~\ref{fig:whiskers_mae}. Despite the high quality of the rules output by CPXR, the method is too selective
and reports only 1.42 rules on average in contrast to \alg{} and MT that find on average 8.81 and 23.92 rules respectively. 
This is also reflected
by the low variance of the reductions compared to other methods. We highlight the large variance of \alg{}+Rfit. 
While it can achieve high positive error reductions, its rule extraction 
is not designed for further filtering, because Rfit reports weak estimators (trees) that become accurate only in combination, for instance, by aggregating their answers or as features for a linear regressor.

All in all, Figure~\ref{fig:tradeoff} suggests that \alg{} offers an interesting trade-off between model complexity and prediction accuracy. This makes it appealing for situations where users need to inspect the correlations that explain the data, or for tuning other methods.

\subsubsection{Runtime Evaluation.}

\begin{figure}
  \centering
  \includegraphics[width=\linewidth]{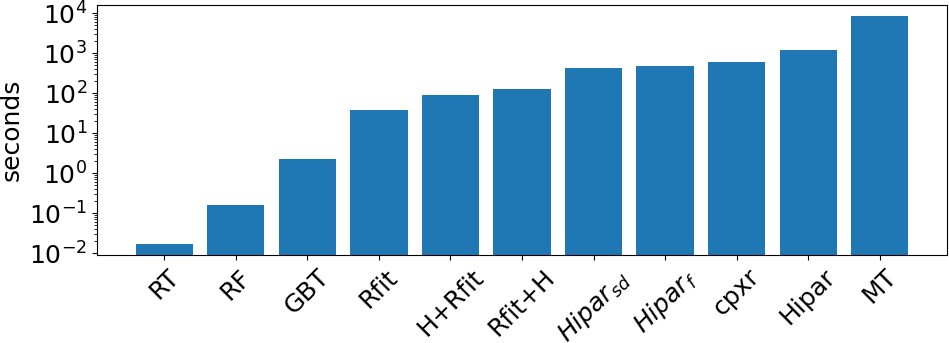}
  \caption{Average runtime of the different methods on the experimental datasets.}
  \label{fig:runtime}
\end{figure}

Figure~\ref{fig:runtime} depicts the average runtime of a fold of cross-validation for the different regression methods. 
We advise the reader to take these results with a grain of salt
because of the heterogeneity of the implementations and the fact that the selection of the parameters (e.g., the minimum support $\theta$) was optimized for error reduction and not runtime. 
RT, GBT, and RF are by far the most performant algorithms partly because 
they count on a highly optimized native scikit-learn implementation.
They are followed by Rfit and the hybrid methods \alg{}+Rfit and Rfit+\alg{}, which combine Rfit with 
\alg{}'s candidate enumeration and rule selection respectively. We observe \alg{} is slower than its variants $\alg_f$ and $\alg_{\textit{sd}}$ because of it adds a more sophisticated rule selection that can take on average 46\% of the total runtime (97\% for \emph{optical}, 0.26\% for \emph{carbon\_nanotubes}).
Finally, we highlight 
that MT is one order of magnitude slower than \alg{}'s 
despite its best-first-search implementation. 
 
\subsection{Anecdotal Evaluation.}
\label{subsec:anecdotal_evaluation}
We illustrate the utility of \alg{} at finding interpretable rules on two use cases used in the evaluation of the EMM approach presented in~\cite{emm-cook-distance}. In this work, the authors introduce the Cook's distance between the coefficients of the default model and the coefficients of the local models as a measure of exceptionality for regions -- referred as subgroups in \cite{emm-cook-distance}. A subgroup with a large Cook's distance is cohesive and its slope vector deviates considerably from the slope vector of the bulk of the data (w.r.t. a target variable). We emphasize that \alg{}'s goal is different from EMM's: The former looks for compact sets of accurate rules, whereas the latter searches for individually exceptional regions. In this spirit, nothing prevents \alg{} from pruning an exceptional region according to EMM if one of its super-regions or sub-regions contributes better to reduce the error. That said, we can neutralize the pruning effect of the selection phase by setting $\omega=0.0$ ($\alg_f$) to make \alg{} output more hybrid rules. This way \alg{} can reproduce the insights of~\cite{emm-cook-distance} for the \emph{wine2} dataset. This dataset consists of 9600 observations derived from 10 years  (1991-2000) of  tasting ratings  reported  in  the online version of the Wine Spectator Magazine for California and Washington red wines. The task is to predict the retail price $y$ of a wine based on features such as its age, production region, grape variety, wine type, etc. We report the best performing set of rules in 5-fold cross-validation. In concordance with~\cite{emm-cook-distance}, this set contains the default rule:
$$
\top \Rightarrow y = -189.69 - 0.0002\times\textit{cases}
+ 2.39\times \textit{score} + 5.08\times\textit{age},
$$
where \emph{score} is the score from the magazine, \emph{age} is the years of aging before commercialization, and \emph{cases} is the number of cases  produced (in thousands). As pointed out in~\cite{emm-cook-distance}, non-varietal wines, i.e., those produced from several grape varieties, tend to have a higher price, and this price is more sensitive to score and age. $\alg_f$ ($\theta=0.05$) found 69 rules including the rule supporting this finding (support 7\%):
\begin{multline}
\textit{variety}=``\textit{non-varietal}" \Rightarrow y = -349.78 - 0.003\times\textit{cases} \\
+ 4.20\times \textit{score} + 7.97\times\textit{age} \nonumber
\end{multline}


\alg{} could also detect the so called \emph{Giffen effect}, observed when, contrary to common sense, the price-demand curve exhibits an upward slope. We observe this phenomenon by running $\alg_f$ on the \emph{giffen} dataset that contains records of the consumption habits of households in the Chinese province of Hunan at different stages of the implementation of a subsidy on staple foodstuffs. The target variable $y$ is the percent change in household consumption of rice, which is predicted via other attributes such as the change in price (\emph{cp}), the household size (\emph{hs}), the income per capita (\emph{ipc}), the calorie consumption per capita (\emph{ccpc}), the share of calories coming from (a) fats (\emph{shf}), and (b) staple foodstuffs (\emph{shs}, \emph{shs2} according to two different definitions), among other indicators. \alg{} finds the default rule:
\begin{multline}
\top \Rightarrow y = 37.27 \textbf{- 0.06} \times \textit{cp} + 1.52\times\textit{hs}  + 0.0004\times{ipc} + 0.003 \times \textit{ccpc} \\ -146.28\times\textit{shf} + 54.13 \times \textit{shs} - 156.78 \times \textit{shs2} \nonumber
\end{multline}
\noindent The negative sign of the coefficient for \emph{cp} suggests no Giffen effect at the global level. As stated in~\cite{emm-cook-distance}, when the subsidy was removed (characterized by the condition \emph{round}=3), the Giffen effect was also not observed in affluent and very poor households. It rather concerned those moderately poor who, despite the surge in the price of rice, increased their consumption at the expense of other sources of calories. Such households can be characterized by intervals in the income and calories per capita (\emph{ipc}, \emph{ccpc}), or by their share of calories from staple foodstuffs (\textit{sh}, \textit{sh2}). This is confirmed by the hybrid rule (support 4\%):
\begin{multline}
\textit{round}=3 \land cpc \in [1898, 2480) \land \textit{sh2} \in [0.7093, \infty) \Rightarrow y = 42.88 \\
\mathbf{+ 1.17}\times \textit{cp} + 1.08\times\textit{hs} - 0.005 \times \textit{ipc} + 0.018 \times \textit{ccpc} \\
-7.42\times\textit{shf} -3.08 \times \textit{shs} - 114.21 \times \textit{shs2} . \nonumber
\end{multline}
\noindent The positive coefficient associated to \emph{cp} shows the Giffen effect for households of moderate calories per capita, whose calories share from staple food is higher than 0.7093. The latter condition aligns with the results of~\cite{emm-cook-distance} that suggested that households with higher values for this variable were more prone to this phenomenon.

\section{Conclusions and Outlook}
We have presented \alg{}, a pattern-aided regression method designed for heterogenous and multimodally distributed
data. \alg{} mines compact sets of accurate hybrid rules thanks to (1) a 
novel hierarchical exploration of the search space of data regions, and (2) a 
selection strategy that optimizes for small sets of rules with joint low prediction error and good coverage. 
\alg{} mines fewer rules than state-of-the-art methods at comparable performance.
As future work, we envision to extend the rule language bias to allow for negated
conditions as in RT and MT, and increase the exhaustivity in the quest for 
accurate hybrid rules. We also envision to parallelize the candidates enumeration phase, and apply other quality criteria and metrics in the search, e.g., the p-values of the linear coefficients. 
As a natural follow-up, we envision to port the notion of hybrid rules to the problem of classification.
\bibliographystyle{ACM-Reference-Format}
\bibliography{references}
\end{document}